\pdfoutput=1
\documentclass[letterpaper, 10 pt, conference]{ieeeconf}

\IEEEoverridecommandlockouts

\usepackage{graphics}
\usepackage{epsfig}
\usepackage{amssymb}

\usepackage{macros}
\title{\LARGE \bf
Tension Tells the Tail: Cable--Obstacle Interaction Sensing\\for Tethered Mobile Robot Navigation
}

\author{Ryuki Shimada, Koki Suzuki, and Genya Ishigami%
\thanks{The authors are with the Space Robotics Group, Department of Mechanical Engineering, Keio University, Yokohama, Kanagawa 223-8522, Japan.
        {\tt\small \{ryukishimada218, curious\}@keio.jp, ishigami@mech.keio.ac.jp}}%
}

\begin{document}

\twocolumn[
\noindent
\copyright~2026 IEEE. Personal use of this material is permitted. Permission from IEEE must be obtained for all other uses, in any current or future media, including reprinting/republishing this material for advertising or promotional purposes, creating new collective works, for resale or redistribution to servers or lists, or reuse of any copyrighted component of this work in other works.\\

\noindent
\textbf{Accepted article:}\\
R. Shimada, K. Suzuki, and G. Ishigami, ``Tension Tells the Tail: Cable--Obstacle Interaction Sensing for Tethered Mobile Robot Navigation,'' \textit{2026 IEEE International Symposium on Safety, Security, and Rescue Robotics (SSRR)}.
]
\thispagestyle{empty}
\pagenumbering{gobble}
\clearpage

\maketitle
\thispagestyle{empty}
\pagestyle{empty}

\begin{abstract}
Tethered mobile robots can sustain long-duration operation through continuous power supply and stable communication via their cables. However, cable--environment interaction causes unintended obstacle displacement, cable snagging or entanglement, and even cable damage, leading to mission failure. To assess the risk of individual cable--obstacle contacts, this study investigates \textit{cable tension as a sensing modality} for estimating whether a contacted obstacle is movable or non-movable. We conducted experiments with a tethered mobile robot and collected tension data while varying robot velocity, contact angle between the obstacle and cable, obstacle mass, and obstacle side-surface material. The results show that obstacle movability is characterized by the initial tension rise and, more importantly, by whether the tension relaxes or remains high after its peak. Based on these observations, we select six features from the tension profiles and train a logistic-regression classifier using 117 labeled trials. The proposed method distinguishes movable and non-movable cases with a balanced accuracy of 95.5\% after 30~cm of post-contact travel. These results demonstrate the potential of cable tension for risk assessment of cable--obstacle interactions and provide a step toward online, cable-aware autonomous navigation.
\end{abstract}

\section{Introduction}\label{sec:introduction}
\label{sec:intro}
Tethered mobile robots are essential platforms for extended operation in hazardous environments. By maintaining a physical connection to a remote base station via cable, the robots receive a continuous power supply and stable communication in environments where wireless transmission is unreliable or impossible, for example, in post-disaster nuclear buildings~\cite{nagatani2011quince} and subterranean spaces~\cite{martz2020surveysubterranean}. The cable itself also serves as a physical lifeline, preventing the robot from falling off steep terrain and providing emergency recovery through cable retrieval. These advantages enable tethered robots to operate safely in diverse and challenging field environments.

\begin{figure}[t]
    \centering
    \includegraphics[width=0.95\linewidth]{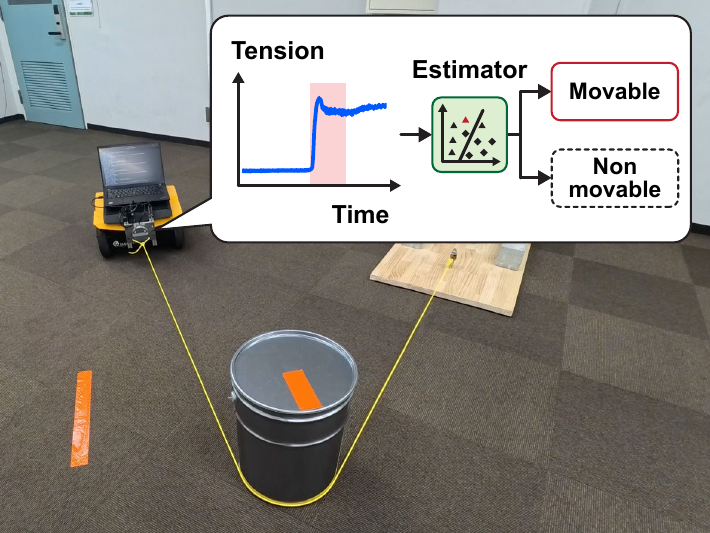}
    \caption{Overview of the cable--obstacle interaction sensing task. As the robot moves, the cable contacts the obstacle and generates tension; the goal is to estimate whether the obstacle is movable or non-movable from the tension measured within a short travel distance after contact.}
    \label{fig:task-overview}
\end{figure}

However, when a tethered robot navigates in confined or unstructured environments, its cable inevitably comes into contact with obstacles (Fig.~\ref{fig:task-overview}), and these interactions can threaten mission completion. Such contacts may cause unintended environmental changes, cable entanglement, cable damage, and ultimately robot immobilization. A notable example occurred during the post-accident investigation of the interior of the Fukushima nuclear power plant in 2011, when the communication cable of the tethered robot Quince~1 became snagged on piping, leading to cable damage and irreversible robot loss~\cite{nagatani2013emergency}. As this case exemplifies, cable--obstacle contact can, in the worst case, render the robot completely disabled and unable to complete its mission.

To address this problem, one research direction has focused on path and motion planning methods that tackle the problem of reduced moving area caused by cable--obstacle interactions under finite cable length. Following the first work on path planning under limited cable length~\cite{xavier1999shortest}, subsequent research~\cite{igarashi2010homotopic} introduced the mathematical concept of homotopy---which distinguishes paths by whether they traverse the left or right of each obstacle---and formalized it mathematically~\cite{kim2014path}. They have been extended to coverage planning~\cite{shnaps2014online}, online exploration~\cite{shapovalov2020exploration}, and motion planning for steep terrain~\cite{abad2011steep}. These approaches, however, share two limitations: they assume obstacles are immovable, and they are indifferent to cable--obstacle contact. More recent work has explicitly addressed cable--obstacle contact avoidance~\cite{shimada2023path,shimada2024tangle}; however, these approaches remain conservative, i.e., avoiding all cable--obstacle interactions. They lack assessment of individual obstacles to determine which contacts are actually dangerous and should be definitely avoided.

\begin{figure}[t]
    \centering
    \includegraphics[width=0.95\linewidth]{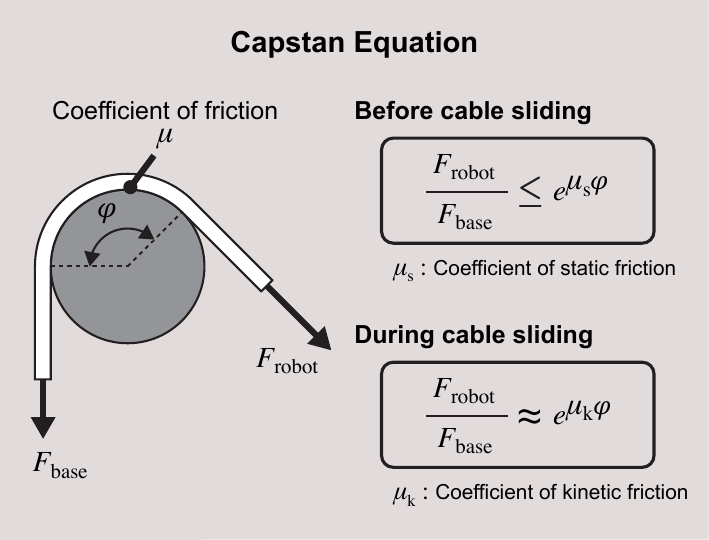}
    \caption{Cable friction around a cylindrical obstacle. Capstan equation governs the tension distribution along the cable--obstacle contact region.}
    \label{fig:capstan}
\end{figure}

In parallel, another research direction has focused on the active exploitation of cable--obstacle interactions to extend robot capabilities. A tethered robotic system called TReX~\cite{mcgarey2016trex} can wrap the cable around natural or artificial objects, enabling traversal of near-vertical slopes. This system also realizes cable-aware SLAM ~\cite{mcgarey2017tslam} by recording contact positions between the cable and pole-like obstacles. Additional work~\cite{page2022robustness} has leveraged the Capstan effect, in which wrapping a cable around an object exponentially amplifies holding force, as illustrated in Fig.~\ref{fig:capstan}. When a large robot equipped with a cable drum becomes immobilized, a small robot at the cable end can wrap the cable around a nearby object such as a tree. The resulting Capstan effect amplifies the holding force, allowing the large robot to achieve self-recovery by reeling the cable. While these approaches successfully extend robot mobility and manipulation capabilities, they lack evaluation to determine which obstacles the robot should contact or wrap the cable around.

In this work, we investigate the use of \textit{cable tension as a sensing modality} for estimating obstacle movability in tethered mobile robot systems. This tension-based sensing approach is particularly valuable in low-visibility environments, where obstacles may be hidden by smoke, darkness, or other objects, because it enables obstacle movability estimation from tension signals alone. We acquire tension data through controlled experiments and develop a machine learning-based classifier that determines whether contacted obstacles are movable or non-movable by analyzing time- and frequency-domain characteristics of the tension signals.

This foundational method addresses a critical gap common to both avoidance and exploitation of cable--obstacle interactions, thereby supporting the real-world deployments of these strategies. Specifically, we contribute:

\begin{enumerate}[label=(\roman*)]
    \item \textbf{An experimental characterization of cable--obstacle interaction behaviors} that identifies the dominant cable--obstacle interaction behaviors and analyzes the influence of robot velocity, contact geometry, and obstacle surface properties on the resulting tension signals.

    \item \textbf{A tension-based obstacle movability estimation method} that classifies movable and non-movable interaction behaviors from tension features within 0.30~m after contact.
\end{enumerate}

\section{Problem Statement}\label{sec:problem-statement}
In this paper, we define the problem as estimating obstacle movability from tension signals measured during cable--obstacle contact (Fig.~\ref{fig:task-overview}). When the cable makes contact with an obstacle, the force sensor mounted at the rear of the robot records the resulting tension profile. This response depends on obstacle properties (mass, shape, ground contact friction, side-surface friction) and interaction parameters (robot velocity, contact geometry). We argue that the robot can distinguish whether the obstacle is displaced or remains in place based on tension signatures measured within a short moving distance after contact.

Here we define two categories of obstacles according to how they behave during cable--obstacle interaction:
\begin{itemize}
    \item \textbf{Movable obstacle:} An obstacle whose position changes as a result of cable–obstacle interaction.
    \item \textbf{Non-movable obstacle:} An obstacle whose position does not change as a result of cable–obstacle interaction. This category also includes obstacles that only rotate in place during cable contact. As a first step toward assessing cable--obstacle interactions, we focus on changes in obstacle position; other effects such as cable snagging are discussed in Section~\ref{subsec:discussion}.
\end{itemize}
Note that \textit{non-movable} does not mean the obstacle is physically immovable or fixed to the environment. Non-movability is not an intrinsic property of the obstacle; it is defined by the interaction behavior, which depends on the specific robot, cable, contact geometry, and other environmental conditions.

The input and output of the system are defined as follows:
\begin{itemize}
    \item \textbf{Input:} Tension signal after contact detection, over a time window scaled to robot velocity.
    \item \textbf{Output:} Binary class (movable or non-movable) with a predicted probability.
\end{itemize}
To ensure that tension variations primarily arise from the cable--obstacle contact, we make the following assumptions:
\begin{itemize}
    \item \textbf{Controlled robot motion:} During each interaction, we command the robot to move straight at constant velocity.
    \item \textbf{Single-obstacle interaction:} Each cable--obstacle interaction involves a single obstacle.
\end{itemize}

\section{Experimental Setup}\label{sec:experimental-setup}
\begin{figure*}[!t]
    \centering
    \includegraphics[width=0.30\textwidth]{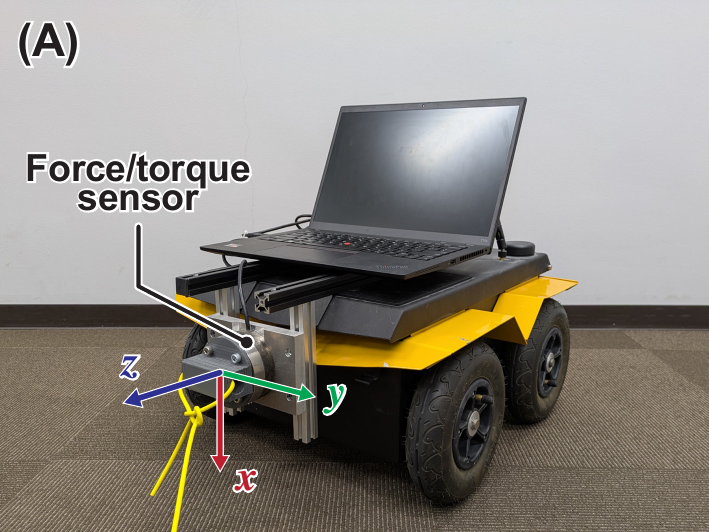}\hfill
    \includegraphics[width=0.30\textwidth]{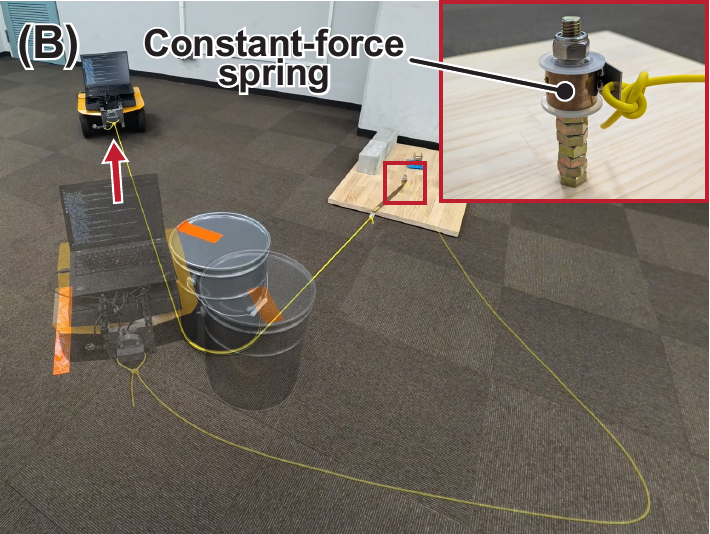}\hfill
    \includegraphics[width=0.30\textwidth]{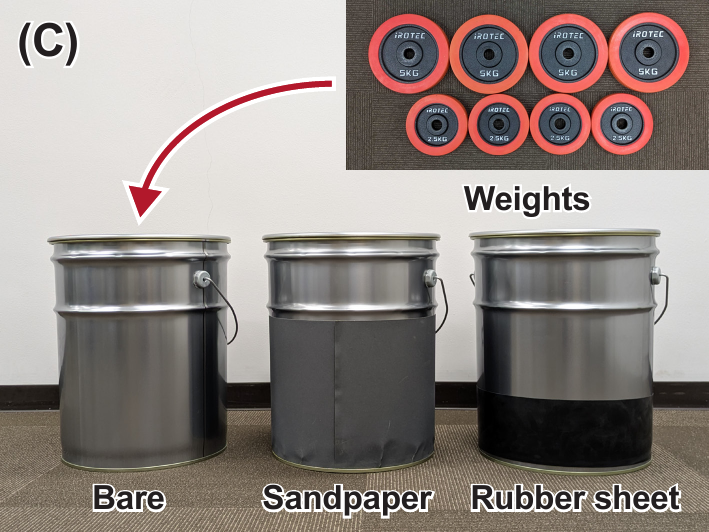}
    \caption{Experimental setup. (A) Jackal robot with a force/torque sensor at its rear; the sensor $z$-axis is aligned with the robot's longitudinal axis. (B) Test field with the cable base, whose constant-force spring (close-up in the red box) keeps the cable pretension constant; the red arrow shows the robot's direction of travel. (C) Steel cans with three side-surface materials (bare steel, sandpaper \#400, and rubber sheet), loaded with weights (top right) to set the obstacle mass.}
    \label{fig:experimental_setup}
\end{figure*}

We designed the experiments to isolate the cable--obstacle interaction as the primary source of tension variation, using the setup shown in Fig.~\ref{fig:experimental_setup}. This section describes the tension measurement system, experimental conditions, and protocol.

\begin{table*}[!t]
\centering
\begin{threeparttable}
\caption{Hardware specifications.}
\label{tab:hardware}
\setlength{\tabcolsep}{5.5pt}
\begin{tabular}{@{}llll@{}}
\toprule
Category   & Model                            & Size                                & Additional specifications \\
\midrule
Robot      & Clearpath Robotics Jackal        & 580 $\times$ 430 $\times$ 250~mm    & Mass 17~kg, Max speed 2.0~m/s \\
Cable      & Blue Robotics Fathom ROV Tether  & $\varnothing$4~mm $\times$ 3~m      & --- \\
Obstacle   & 20~L steel can          & $\varnothing$306~mm $\times$ 367~mm & Mass 1.25~kg (empty) \\
Force/torque sensor & Leptrino PFS080YS102U6           & $\varnothing$80~mm $\times$ 35~mm   & Max$^{*}$ $\pm$1000~N ($F_z$), Resolution 0.25~N, Frequency $\leq$1.2~kHz \\
Constant-force spring & MISUMI CFS 1.4          & $\varnothing$25.6~mm                & Pretension 13.73~N, Stroke 1000~mm \\
\bottomrule
\end{tabular}
\begin{tablenotes}[flushright]
\footnotesize
\item[*] Denotes rated capacity, not allowable overload limit.
\end{tablenotes}
\end{threeparttable}
\end{table*}

\subsection{Cable Tension Measurement System}
Our system consists of three components: a mobile robot mounting a force/torque sensor at the rear side, a cable, and a cable base with a constant-force spring. The exact specification of hardware is summarized in Table~\ref{tab:hardware}.

\textbf{Robot with force/torque sensor.}
The robot platform is a Clearpath Robotics Jackal robot with a Leptrino six-axis force/torque sensor mounted between the cable end and the robot (Fig.~\ref{fig:experimental_setup}A). The sensor $z$-axis is aligned with the robot's longitudinal axis, along which the robot pulls the cable. We set the sampling rate to 500~Hz to capture both the slow tension build-up at contact onset and higher-frequency oscillations expected from stick--slip behavior.

\textbf{Cable.}
We use a cable for underwater robots, following a prior tethered system TReX~\cite{mcgarey2016trex}. This cable exhibits negligible stretch under the tension ranges in our experiments, ensuring that tension measurements reflect only cable--obstacle interactions. Its low bending stiffness allows conformance to obstacle surfaces during contact.

\textbf{Cable base with constant-force spring.}
The cable base, which anchors the cable end to the ground, uses a passive unreeling mechanism with a constant-force spring (Fig.~\ref{fig:experimental_setup}B) that keeps the cable pretension constant while letting the cable extend as the robot moves. We use this passive mechanism instead of an active cable drum with closed-loop tension control to avoid introducing actuator dynamics into the measured signals.

\subsection{Experimental Conditions and Protocol}
\label{subsec:conditions_protocol}

\begin{figure}[t]
    \centering
    \includegraphics[width=0.95\linewidth]{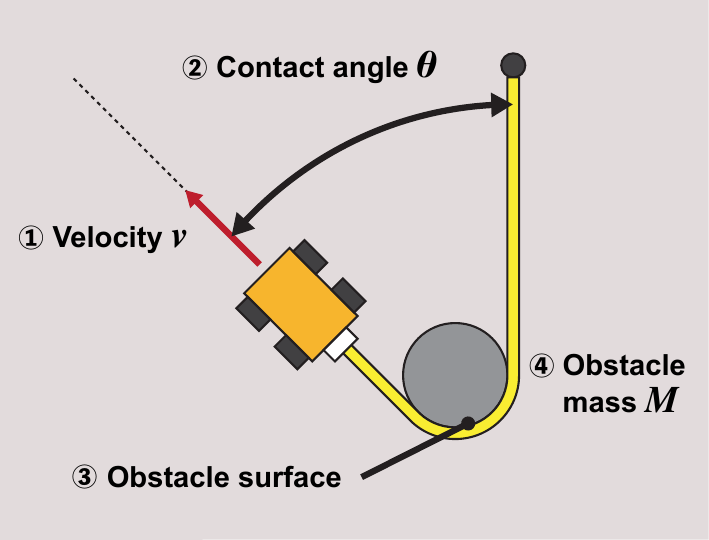}
    \caption{Experimental conditions: (1) robot velocity, (2) contact angle, (3) side-surface material, and (4) obstacle mass.}
    \label{fig:experimental_conditions}
\end{figure}

\textbf{Experimental conditions.}
We vary the following four parameters while keeping the ground condition (tile carpet) and obstacle shape (cylindrical) fixed (Fig.~\ref{fig:experimental_conditions}):

\begin{itemize}
    \item \textbf{Robot velocity $v$:} 0.05, 0.10, and 0.20~m/s.
    \item \textbf{Contact angle $\theta$:} the angle between the line from the cable base to the obstacle and the direction from the obstacle to the robot: $45$, $90$, and $135^\circ$.
    \item \textbf{Obstacle mass:} 6.25, 16.25, and 31.25~kg, obtained by adding weights inside the steel can (1.25~kg) (Fig.~\ref{fig:experimental_setup}C).
    \item \textbf{Side-surface material:} bare steel, sandpaper \#400, and rubber sheet; sandpaper and rubber are tested only at $\theta=45^\circ$ with the 16.25 and 31.25~kg obstacles.
\end{itemize}

\textbf{Protocol.}
For each condition combination, we conduct three trials (117 in total). In each trial, we place the steel can at a grid-aligned position on the floor (50~cm $\times$ 50~cm tiles) and set the robot at an initial pose whose heading is adjusted to the target contact angle $\theta$, with the cable in its initial configuration (ghosted overlay in Fig.~\ref{fig:experimental_setup}B). We then command the robot to move straight at a constant velocity for a duration proportional to the velocity, so that the recording captures the post-contact dynamics at all speeds. Tension and video are recorded throughout each trial.

\begin{figure*}[!t]
    \centering
    \includegraphics[width=0.30\textwidth]{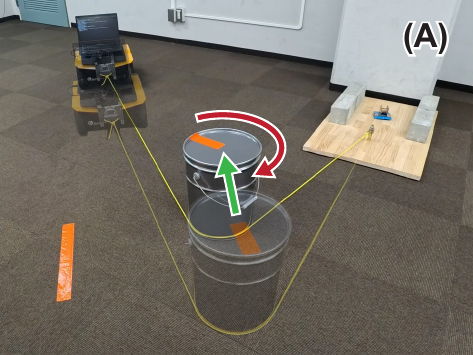}\hfill
    \includegraphics[width=0.30\textwidth]{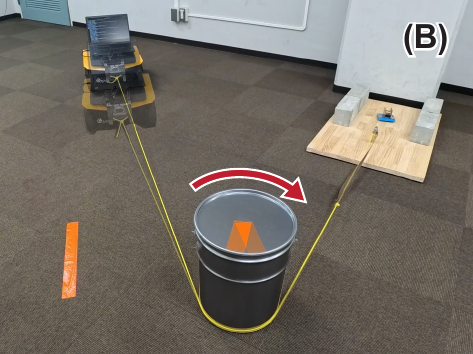}\hfill
    \includegraphics[width=0.30\textwidth]{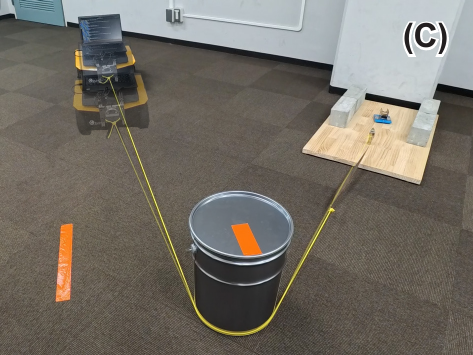}
    \caption{Representative examples of cable--obstacle interaction behaviors. (A) Rotation and translation (movable): the obstacle rotates (red arrow) and translates (green arrow) on the ground. (B) Rotation only (non-movable): the obstacle rotates in place without translating. (C) Stationary (non-movable): the obstacle stays in place while the cable slides along its side surface.}
    \label{fig:sample-outcomes}
    \vspace{-7pt}
\end{figure*}

\textbf{Ground-truth labeling of obstacle behavior.}
We manually assign labels by visual inspection of recorded videos, distinguishing three categories: \textit{movable} (visible positional displacement; Fig.~\ref{fig:sample-outcomes}A), \textit{rotation-only} (rotational motion without translation; Fig.~\ref{fig:sample-outcomes}B), and \textit{stationary} (no obstacle motion and cable sliding; Fig.~\ref{fig:sample-outcomes}C). For classification, the latter two are labeled as \textit{non-movable}, as defined in Section~\ref{sec:problem-statement}.

\section{Experimental Results}~\label{subsec:exp-results}
We characterize cable--obstacle interactions by analyzing the tension signal properties that are useful for movability classification. As the cable tension, we use $F_z$ measured in the longitudinal direction of the robot (Fig.~\ref{fig:experimental_setup}A): under all tested geometries, the contact-induced force deviates from the $z$-axis by $9^\circ$ (median; at most $14^\circ$), and thus $F_z$ captures at least 97\% of the contact-induced change in the force magnitude $\|(F_x, F_y, F_z)\|$. We denote the change in $F_z$ from its pre-contact level by $\Delta F_z$. Our observations, detailed below, lead to three guidelines for feature design:
\begin{itemize}
\item \textbf{Capture the initial tension rise}, which reflects how the cable-to-obstacle force approaches the static friction limit with the ground.
\item \textbf{Capture the post-peak transition}, which reflects whether the tension relaxes as the obstacle starts sliding on the ground (movable) or stays high as the cable slides along the side surface (non-movable).
\item \textbf{Capture the late-phase residual tension and fluctuation}, which reflect whether the tension fluctuates with obstacle--ground stick--slip (movable) or remains high with cable sliding (non-movable).
\end{itemize}

\begin{figure*}
    \centering
    \includegraphics[width=\linewidth]{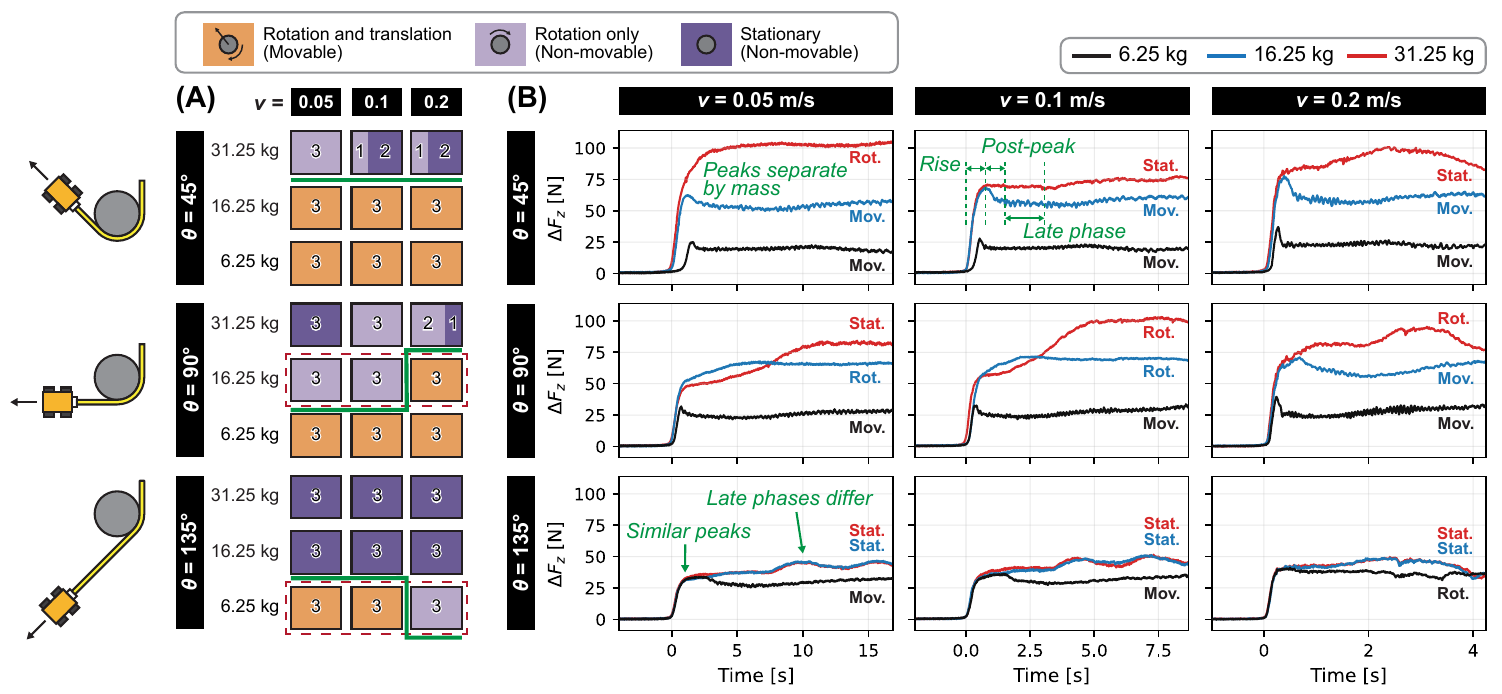}
    \caption{Behaviors and tension profiles on the bare surface. (A) Interaction behavior in each condition; numbers denote trial counts. Green thick lines mark the movable/non-movable boundary, and red dashed boxes mark the conditions in which velocity changes the behavior. (B) Representative $\Delta F_z$ profile of the majority behavior in each condition. Time axes cover similar robot travel distances, and green arrows at $\theta=45^\circ$ and $v=0.10$~m/s mark the three phases used for feature design.}
    \label{fig:grid-3x3}
    \vspace{-7pt}
\end{figure*}

\begin{figure}[t]
    \centering
    \includegraphics[width=0.80\linewidth]{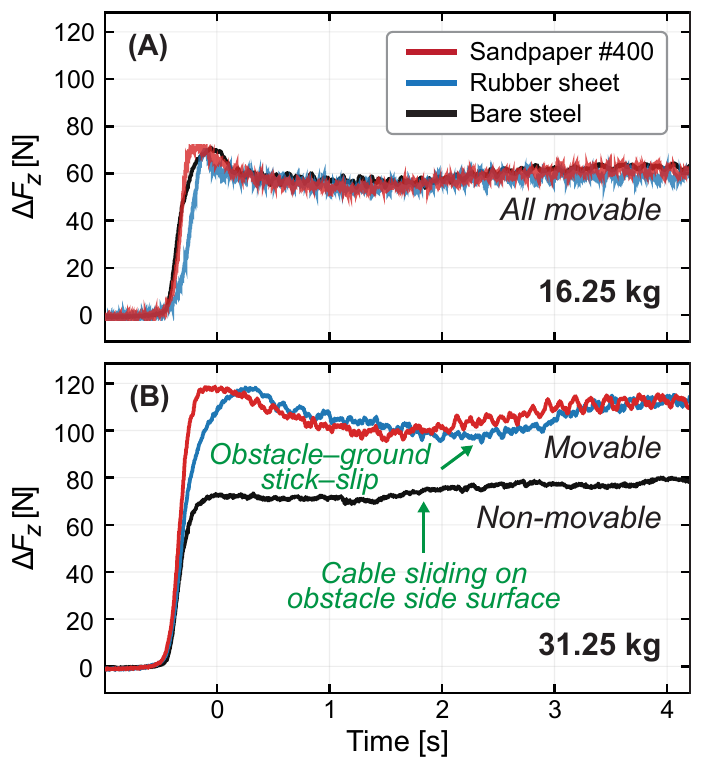}
    \caption{Mean tension profiles over three trials for different obstacle side surfaces at $v=0.10$~m/s and $\theta=45^\circ$. (A) 16.25~kg. (B) 31.25~kg.}
    \label{fig:exp3-friction}
\end{figure}

Across all conditions, obstacle mass primarily governs the movable/non-movable behavior, while robot velocity, contact angle, and obstacle side-surface material shift the resulting boundary (Figs.~\ref{fig:grid-3x3}A and~\ref{fig:exp3-friction}). For each factor, we show how the boundary shifts, how the tension signal changes, and how these changes provide cues for movability estimation.

\subsection{Effect of Robot Velocity}
Robot velocity changes the interaction behavior only in a few conditions near the movable/non-movable boundary (Fig.~\ref{fig:grid-3x3}A, dashed boxes). The 6.25~kg obstacle is movable and the 31.25~kg obstacle is non-movable in nearly all conditions. Near the boundary, the 16.25~kg obstacle becomes movable only at $v=0.20$~m/s at $\theta=90^\circ$, and the 6.25~kg obstacle becomes non-movable only at $v=0.20$~m/s at $\theta=135^\circ$.

Velocity mainly compresses the tension profiles in time without changing their behavior-dependent pattern (Fig.~\ref{fig:grid-3x3}B). At $\theta=45^\circ$, movable cases show a sharp rise and a transient peak, followed by kinetic friction and stick--slip, whereas non-movable cases show cable sliding, at all velocities. Therefore, the initial rise and the post-peak behavior remain useful cues for movability estimation across speeds. This compression in time also suggests that features should be extracted from the same interaction phase, rather than from a fixed time interval.

\subsection{Effect of Contact Angle}
A larger contact angle shifts the movable/non-movable boundary toward lighter obstacles (Fig.~\ref{fig:grid-3x3}A, thick lines). At $\theta=45^\circ$, the 16.25~kg obstacle is movable, while the 31.25~kg obstacle is non-movable. At $\theta=90^\circ$, the 16.25~kg obstacle is near the boundary. At $\theta=135^\circ$, both the 16.25~kg and 31.25~kg obstacles are non-movable, and even the 6.25~kg obstacle becomes non-movable at the highest speed. This shift can be explained by the smaller effective wrap angle of the cable around the obstacle ($\varphi$ in Fig.~\ref{fig:capstan}), which shortens the contact area and weakens force transmission.\looseness=-1

A larger contact angle reduces the mass-dependent difference in the peak force but preserves the late-phase difference (Fig.~\ref{fig:grid-3x3}B). At $\theta=45^\circ$, the profiles for different masses are clearly separated, and the peak force helps distinguish movable and non-movable cases. These differences become smaller at $\theta=90^\circ$ and even smaller at $\theta=135^\circ$, so peak force alone becomes less reliable. However, the late-phase behavior remains different even at $\theta=135^\circ$: movable cases keep kinetic friction and stick--slip, whereas non-movable cases show cable sliding. Therefore, contact angle makes peak-based estimation less reliable, but it preserves the usefulness of late-phase residual tension and fluctuation as cues for movability estimation.

\subsection{Effect of Obstacle Side-surface Material}
A higher-friction side surface shifts the movable/non-movable boundary toward heavier obstacles (Fig.~\ref{fig:exp3-friction}). The 16.25~kg obstacle is movable on all materials, with largely overlapping late-phase profiles (Fig.~\ref{fig:exp3-friction}A), because the late-phase tension is then governed by obstacle--ground stick--slip rather than by the side surface. In contrast, the 31.25~kg obstacle is non-movable on bare steel, but translates on sandpaper \#400 and rubber (Fig.~\ref{fig:exp3-friction}B). This boundary shift is consistent with the capstan equation (Fig.~\ref{fig:capstan}): a larger cable--obstacle friction coefficient $\mu$ allows the cable to transmit larger force before slipping.

The late-phase tension profiles also reflect this shift (Fig.~\ref{fig:exp3-friction}B): bare steel keeps the cable-sliding profile, while high-friction surfaces show kinetic friction and stick--slip, again supporting late-phase residual tension and fluctuation as robust cues. Within the high-friction surfaces, sandpaper \#400 rises to its peak more steeply than rubber (Fig.~\ref{fig:exp3-friction}). This difference may reflect different friction mechanisms: mechanical interlocking for sandpaper and adhesion-related friction for rubber.  We do not further analyze this difference because it does not affect the movability estimation.

\section{Cable-Obstacle Interaction Estimation}
\begin{figure*}[t]
    \centering
    \includegraphics[width=0.99\linewidth]{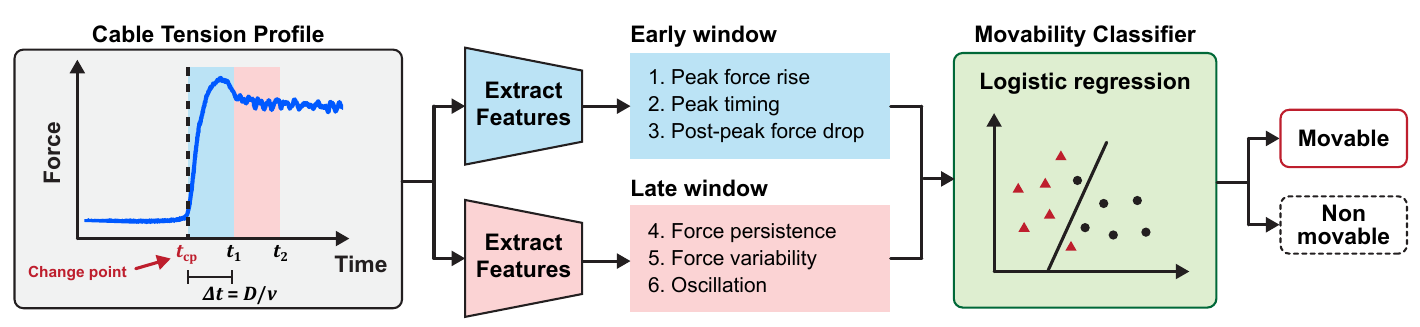}
    \caption{Pipeline for movability classification. For each tension signal, we detect the change point $t_{\mathrm{cp}}$ and extract three features from each of two post-contact windows with equal duration. The resulting six features are used in a logistic regression classifier to classify the contacted obstacle as movable or non-movable.}
    \label{fig:pipeline}
\end{figure*}

\begin{table*}[t]
  \centering
  \caption{Six features extracted from cable tension $F_z$. Definitions are relative to the contact onset $t_\mathrm{cp}$, where \emph{early} and \emph{late} denote two consecutive windows after $t_\mathrm{cp}$, each of length $\Delta t = D/v$. The baseline force $f_\mathrm{base}$ is the mean of $F_z$ over $[t_\mathrm{cp}-\Delta t, t_\mathrm{cp})$; the peak force $f_\mathrm{peak}$ and the peak time $t_\mathrm{peak}$ are the maximum of $F_z$ in the early window and its time from $t_\mathrm{cp}$; and the end force $f_\mathrm{end}$ is the last sample of the early window.}
  \label{tab:features}
  \footnotesize
  \renewcommand{\arraystretch}{1.25}
  \begin{tabular}{
    >{\centering\arraybackslash}m{0.06\textwidth}
    >{\centering\arraybackslash}m{0.03\textwidth}
    >{\raggedright\arraybackslash}m{0.24\textwidth}
    >{\raggedright\arraybackslash}m{0.18\textwidth}
    >{\raggedright\arraybackslash}m{0.30\textwidth}
    >{\centering\arraybackslash}m{0.04\textwidth}
  }
    \toprule
    Window & No. & Feature name & Notation & Definition & Unit \\
    \midrule
    \multirow{3}{*}{Early} & 1 & Peak force rise & \texttt{rise\_to\_peak} & $f_\mathrm{peak} - f_\mathrm{base}$ & N \\
     & 2 & Normalized peak timing & \texttt{peak\_time\_ratio} & $t_\mathrm{peak} / $$\Delta t$ & -- \\
     & 3 & Post-peak force drop & \texttt{drop\_ratio} & $(f_\mathrm{peak} - f_\mathrm{end}) / $$(f_\mathrm{peak} - f_\mathrm{base})$ & -- \\
    \midrule
    \multirow{3}{*}{Late} & 4 & Late-to-peak force ratio & \texttt{late\_mean\_over\_peak} & $(\bar{F}_z^\mathrm{late} - f_\mathrm{base}) / (f_\mathrm{peak} - f_\mathrm{base})$ & -- \\
     & 5 & Late-phase coefficient of variation & \texttt{wlate\_cv} & $\sigma_z^\mathrm{late} / (\bar{F}_z^\mathrm{late} - f_\mathrm{base})$ & -- \\
     & 6 & Late-phase zero-crossing density & \texttt{wlate\_zc\_detrended} & Sign changes per meter after linear detrending & 1/m \\
    \bottomrule
  \end{tabular}
\end{table*}

We formulate cable--obstacle interaction estimation as a binary movability classification problem. Fig.~\ref{fig:pipeline} illustrates the overall pipeline, from change-point detection and feature extraction to movable/non-movable classification.

\subsection{Dataset}
The dataset contains 117 trials and covers three robot velocities, three contact angles, three obstacle masses, and three side-surface materials, as described in Section~\ref{subsec:conditions_protocol}. The labels consist of 71 movable and 46 non-movable trials. All movable trials involve obstacle \textit{rotation and translation}, whereas the non-movable class includes both \textit{stationary} (26/46) and \textit{rotation-only} (20/46) cases.

\subsection{Feature Engineering}\label{subsec:feature-engineering}
\textbf{Change-point detection.}
We detect the change-point time $t_\mathrm{cp}$, which marks contact onset (Fig.~\ref{fig:pipeline}), offline as the first time at which the moving average of $F_z$ over a window $\tau$ stays above the threshold $F_0 + \alpha\,(F_{95} - F_0)$ for $\tau$. Here, $F_0$ is the mean of the moving-averaged $F_z$ in the initial part of the recording time, and $F_{95}$ is its 95th percentile over the whole recording time. We empirically set $\tau = 0.35$~s and $\alpha = 0.08$, and compute $F_0$ over the first 12\% of the recording time (at most 2~s).

\textbf{Time windows.}
We extract features from two post-contact windows. To compare trials with different robot velocities, we define each window by a travel distance $D$ after contact, rather than by a fixed time length. The corresponding time length is $\Delta t = D/v$, where $v$ is the robot velocity. The early and late windows are then defined as
$W_\mathrm{early} = [t_\mathrm{cp}, t_\mathrm{cp}+\Delta t)$ and
$W_\mathrm{late} = [t_\mathrm{cp}+\Delta t, t_\mathrm{cp}+2\Delta t)$.
Features are extracted separately from these two windows.

Here, $D$ is a user-defined parameter that controls how much post-contact dynamics each window includes: a small $D$ covers mainly the initial tension rise, whereas a larger $D$ also includes the post-peak force drop and the late-phase response (Fig.~\ref{fig:d_coverage}).

\textbf{Feature extraction.}
We extract six features (Table~\ref{tab:features}), chosen based on the observations in Section~\ref{subsec:exp-results}, from the $F_z$ signal. From $W_\mathrm{early}$, we extract features related to the transient response: the initial force rise, peak timing, and post-peak drop. From $W_\mathrm{late}$, we extract features related to the late-phase response: the late-to-peak force ratio, late-phase coefficient of variation, and late-phase zero-crossing density. Before extraction, $F_z$ is low-pass filtered with a fourth-order zero-phase Butterworth filter at 25~Hz. Above this frequency, the late-window tension contains only noise comparable to that before contact, so the filter removes noise without losing late-phase information. The contribution of each feature and the choice of $D$ are evaluated in Section~\ref{subsec:results}.

\subsection{Classification Algorithm}
\begin{table*}[t]
  \centering
  \caption{Drop-one feature ablation across prediction distances $D$ (stratified 5-fold cross-validation). Each value is the change in mean balanced accuracy when that feature is dropped from the full six-feature set, $\Delta\overline{\mathrm{BA}} = \overline{\mathrm{BA}}_\mathrm{drop} - \overline{\mathrm{BA}}_\mathrm{full}$; more negative values indicate greater importance. Bold marks the largest negative $\Delta\overline{\mathrm{BA}}$ in each column.}
  \label{tab:ablation}
  \footnotesize
  \begin{tabular}{lrrrrrr}
    \toprule
    Dropped Feature
      & $D=0.05$ m & $D=0.10$ m & $D=0.15$ m
      & $D=0.20$ m & $D=0.25$ m & $D=0.30$ m \\
    (post-contact travel $2D$)
      & (0.10 m) & (0.20 m) & (0.30 m)
      & (0.40 m) & (0.50 m) & (0.60 m) \\
    \midrule
    \texttt{rise\_to\_peak}
      & $-0.006$ & $-0.061$ & $-0.053$
      & $\mathbf{-0.080}$ & $\mathbf{-0.044}$ & $\mathbf{-0.024}$ \\
    \texttt{peak\_time\_ratio}
      & $\phantom{+}0.000$ & $-0.011$ & $+0.007$
      & $\phantom{+}0.000$ & $+0.004$ & $\phantom{+}0.000$ \\
    \texttt{drop\_ratio}
      & $\phantom{+}0.000$ & $\phantom{+}0.000$ & $+0.007$
      & $-0.004$ & $\phantom{+}0.000$ & $-0.010$ \\
    \texttt{late\_mean\_over\_peak}
      & $\phantom{+}0.000$ & $\mathbf{-0.083}$ & $\mathbf{-0.076}$
      & $-0.046$ & $-0.021$ & $\phantom{+}0.000$ \\
    \texttt{wlate\_cv}
      & $\mathbf{-0.068}$ & $-0.007$ & $-0.014$
      & $+0.007$ & $\phantom{+}0.000$ & $\phantom{+}0.000$ \\
    \texttt{wlate\_zc\_detrended}
      & $-0.009$ & $+0.007$ & $+0.014$
      & $\phantom{+}0.000$ & $+0.004$ & $\phantom{+}0.000$ \\
    \midrule
    Full set ($\overline{\mathrm{BA}} \pm \sigma$)
      & $0.773 \pm 0.071$ & $0.912 \pm 0.039$ & $0.955 \pm 0.041$
      & $0.969 \pm 0.028$ & $0.965 \pm 0.033$ & $0.976 \pm 0.020$ \\
    \bottomrule
  \end{tabular}
\end{table*}

\textbf{Classifier.}
We use logistic regression with $\ell_2$ regularization and set the regularization parameter to $C=1.0$. To account for the label imbalance, we use balanced class weights. All input features are standardized to zero mean and unit variance before training. We do not tune hyperparameters.

\textbf{Evaluation protocol.}
We use stratified 5-fold cross-validation because the dataset is modest in size. The primary metric is \textit{balanced accuracy}, the average recall over the two classes: $\mathrm{BA} = [\mathrm{TP}/(\mathrm{TP}+\mathrm{FN}) + \mathrm{TN}/(\mathrm{TN}+\mathrm{FP})]/2$.
We report the mean and standard deviation over the five folds as $\overline{\mathrm{BA}} \pm \sigma$. The confusion matrix is computed from out-of-fold predictions over all trials. Performance is evaluated only with cross-validation results; the classifier is retrained on the full dataset only for decision-boundary visualization.

\subsection{Classification Results}\label{subsec:results}
We first analyze the effect of $D$ and feature importance, and then the results at the selected $D$.

\textbf{Effect of prediction distance $D$.}
Classification performance improves with $D$ and then saturates (bottom row of Table~\ref{tab:ablation}). The balanced accuracy reaches $0.955 \pm 0.041$ at $D=0.15$~m and $0.976 \pm 0.020$ at $D=0.30$~m, the highest among the tested values. However, the improvement after $D=0.15$~m is small compared with the additional post-contact travel required before classification. Later observations may also include condition-dependent variations, such as stick--slip timing and cable sliding. We therefore use $D=0.15$~m as a practical balance between accuracy and travel distance. This choice corresponds to a total post-contact travel of $2D=0.30$~m per estimation.

\begin{figure}[t]
    \centering
        \includegraphics[width=0.95\linewidth]{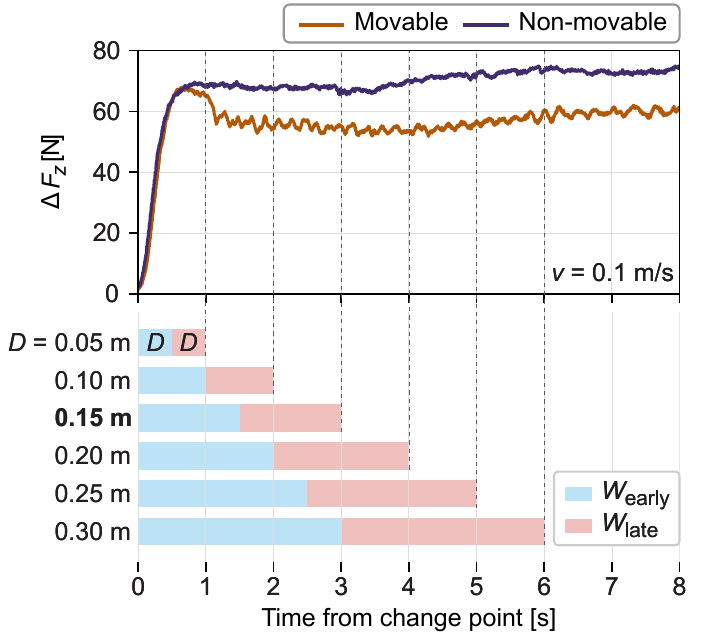}
    \caption{Early and late feature-extraction windows for different travel distances $D$ at $v=0.10$~m/s, where each window has duration $\Delta t = D/v$. Bold marks the selected $D$.}
    \label{fig:d_coverage}
\end{figure}

\textbf{Feature importance across prediction distances.}
The drop-one ablation in Table~\ref{tab:ablation} shows that the important features change with the prediction distance $D$. At the shortest distance ($D=0.05$~m), removing \texttt{wlate\_cv} causes the largest performance drop, suggesting that short-window fluctuation is useful when late-phase information is still limited. In the middle range ($D=0.10$--$0.15$~m), \texttt{late\_mean\_over\_peak} becomes dominant, with large drops of $-0.083$ and $-0.076$. At $D=0.20$--$0.30$~m, \texttt{rise\_to\_peak} becomes dominant. These results support the physical cues identified in Section~\ref{subsec:exp-results}: the initial tension rise and late-phase tension behavior are the main cues for movability estimation, and their importance depends on how much post-contact dynamics the windows cover (Fig.~\ref{fig:d_coverage}).

\begin{figure}[t]
    \centering
        \includegraphics[width=\linewidth]{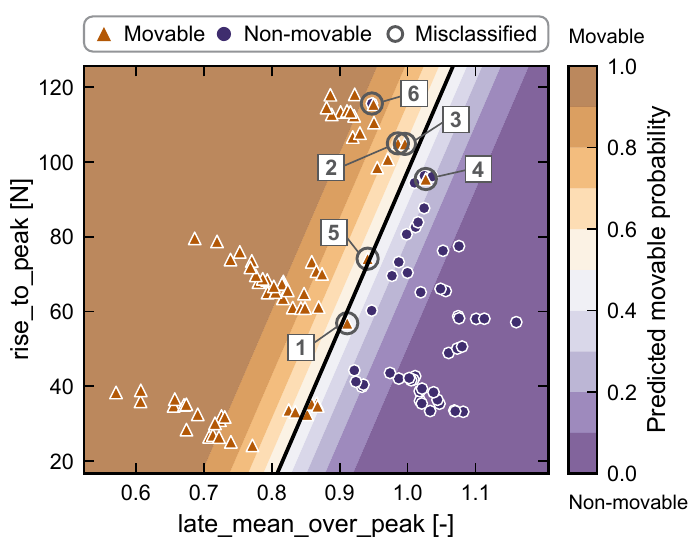}
        \caption{Logistic-regression decision slice at $D=0.15$~m, with the remaining four features fixed at their dataset medians. The solid line is the decision boundary, and circles denote misclassified trials, numbered as in Section~\ref{subsec:results}.}
    \label{fig:decision-slices}
\end{figure}

\textbf{Performance and failure analysis at $D=0.15$~m.}
The decision slice in Fig.~\ref{fig:decision-slices} shows that the classifier separates cases mainly along \texttt{late\_mean\_over\_peak}, that is, by whether the late-phase tension falls below the peak or stays near it, consistent with the ablation results. The out-of-fold confusion matrix, with movable treated as the positive class, gives TP~=~66, TN~=~45, FP~=~1, and FN~=~5.
In all six errors (circles in Fig.~\ref{fig:decision-slices}), the tension resembles that of the other class for one of three reasons:
\begin{itemize}
    \item \textbf{Error (1), false negative:} A movable trial predicted as non-movable (16.25~kg, rubber, $v=0.05$~m/s). The obstacle rotates and translates simultaneously, but the post-peak relaxation is small at the slowest velocity.
    \item \textbf{Errors (2)--(5), false negatives:} Movable trials predicted as non-movable (31.25~kg on sandpaper \#400 (2, 3) and on rubber (4) at $v=0.05$~m/s; 16.25~kg on bare steel at $v=0.20$~m/s (5)). The obstacle first rotates slightly and then translates while rotating, so the tension stays high during the initial rotation.
    \item \looseness=-1 \textbf{Error (6), false positive:} The single non-movable trial predicted as movable (31.25~kg, sandpaper \#400, $v=0.20$~m/s). The obstacle rotates largely in place before the cable slides along it; the resulting high rise and slight late-phase decrease resemble movable trials.
\end{itemize}

\begin{figure}[t]
    \centering
    \includegraphics[width=0.95\linewidth]{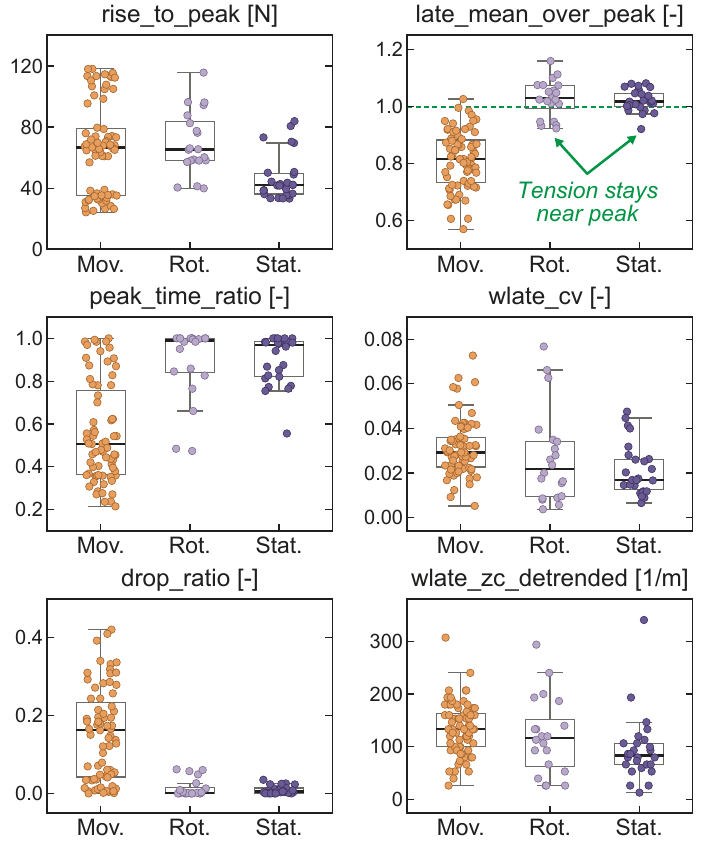}
    \caption{Distributions of the six features for movable (Mov., $n=71$), rotation-only (Rot., $n=20$), and stationary (Stat., $n=26$) interactions at $D=0.15$~m. The green dashed line marks a late-phase tension equal to the peak rise.}
    \label{fig:three-class}
\end{figure}

\subsection{Discussion}\label{subsec:discussion}
We examine how rotation-only interactions, which we treat as non-movable, compare with movable and stationary interactions in the tension features, and then discuss the limitations of this study.

\textbf{Tension features of rotation-only interactions.} Fig.~\ref{fig:three-class} compares the six features among the three observed behaviors at the selected $D$ (0.15~m). Compared with movable cases, rotation-only cases are clearly separated in \texttt{late\_mean\_over\_peak}, \texttt{peak\_time\_ratio}, and \texttt{drop\_ratio}; the tension stays near the peak, as in other non-movable interactions. Compared with stationary cases, rotation-only cases overlap in all features except \texttt{rise\_to\_peak}, which is lower for stationary cases, but 18/26 of them (vs.\ 3/20 rotation-only cases) occur at $\theta=135^\circ$ (Fig.~\ref{fig:grid-3x3}A), where \texttt{rise\_to\_peak} is low for all behaviors. Distinguishing rotation from stationary behavior therefore requires cues beyond the current tension features.

\textbf{Limitations and future directions.}
We identify three limitations of the current study and outline future directions:
\begin{itemize}
    \item \textbf{Environmental diversity:} The controlled setup isolates the essential interaction phenomena, but validation across diverse terrains and obstacle shapes (e.g., cuboid, polygonal prism) remains future work.
    \item \textbf{Stationary non-sliding contact:} The dataset lacks cases where a heavy obstacle stays stationary without cable sliding; probing this regime requires surfaces or geometries that induce cable snagging.
    \item \textbf{Rotation-induced contact:} Obstacle rotation, which occurs in both movable and rotation-only cases, can bring a new side surface into contact with the cable and increase the risk of snagging or entanglement; assessing this risk is left for future work.
\end{itemize}

\section{Conclusion}\label{sec:conclusion}
This paper investigated cable tension as a sensing modality for estimating obstacle movability in tethered mobile robots. Through controlled cable--obstacle interaction experiments, we showed that movable and non-movable behaviors are characterized by the initial tension rise and, more importantly, by the post-peak evolution of the tension profile. Based on these observations, we designed tension-profile features, and a logistic-regression classifier trained on 117 trials achieved a balanced accuracy of 95.5\% after 0.30~m of post-contact travel. These results suggest that cable tension can be useful for online assessment of cable--obstacle interactions that could help a tethered robot decide whether to continue navigation or retreat to resolve contact. Future work will address misclassifications where in-place rotation of the obstacle blurred the tension difference between movable and non-movable obstacles, by extending the tension features or adding a sensing modality that captures such rotation.

\section*{ACKNOWLEDGMENT}
This work was supported by JSPS KAKENHI Grant Number JP26KJ1952.
The authors would like to thank Haruka Nakamura for her support with the early-stage experimental development of this project, and Kosei Yoshida for assistance with the fabrication of the experimental setup.

\bibliography{references}

\begin{thebibliography}{10}

\bibitem{nagatani2011quince}
K.~Nagatani, S.~Kiribayashi, Y.~Okada, S.~Tadokoro, T.~Nishimura, T.~Yoshida,
  E.~Koyanagi, and Y.~Hada, ``Redesign of rescue mobile robot {Quince},'' in
  {\em Proc. IEEE Int. Symp. Safety, Secur., Rescue Robot. (SSRR)}, pp.~13--18,
  IEEE, 2011.

\bibitem{martz2020surveysubterranean}
J.~Martz, W.~Al-Sabban, and R.~N. Smith, ``Survey of unmanned subterranean
  exploration, navigation, and localisation,'' {\em IET Cyber-Syst. Robot.},
  vol.~2, no.~1, pp.~1--13, 2020.

\bibitem{nagatani2013emergency}
K.~Nagatani, S.~Kiribayashi, Y.~Okada, K.~Otake, K.~Yoshida, S.~Tadokoro,
  T.~Nishimura, T.~Yoshida, E.~Koyanagi, M.~Fukushima, {\em et~al.},
  ``Emergency response to the nuclear accident at the {Fukushima Daiichi}
  nuclear power plants using mobile rescue robots,'' {\em J. Field Robot.},
  vol.~30, no.~1, pp.~44--63, 2013.

\bibitem{xavier1999shortest}
P.~G. Xavier, ``Shortest path planning for a tethered robot or an anchored
  cable,'' in {\em Proc. IEEE Int. Conf. Robot. Autom. (ICRA)}, vol.~2,
  pp.~1011--1017, IEEE, 1999.

\bibitem{igarashi2010homotopic}
T.~Igarashi and M.~Stilman, ``Homotopic path planning on manifolds for cabled
  mobile robots,'' in {\em Proc. Workshop Algorithmic Found. Robot. (WAFR)},
  pp.~1--18, Springer, 2010.

\bibitem{kim2014path}
S.~Kim, S.~Bhattacharya, and V.~Kumar, ``Path planning for a tethered mobile
  robot,'' in {\em Proc. IEEE Int. Conf. Robot. Autom. (ICRA)}, pp.~1132--1139,
  IEEE, 2014.

\bibitem{shnaps2014online}
I.~Shnaps and E.~Rimon, ``Online coverage by a tethered autonomous mobile robot
  in planar unknown environments,'' {\em IEEE Trans. Robot.}, vol.~30, no.~4,
  pp.~966--974, 2014.

\bibitem{shapovalov2020exploration}
D.~Shapovalov and G.~A. Pereira, ``Exploration of unknown environments with a
  tethered mobile robot,'' in {\em Proc. IEEE/RSJ Int. Conf. Intell. Robots
  Syst. (IROS)}, pp.~6826--6831, IEEE, 2020.

\bibitem{abad2011steep}
P.~Abad-Manterola, I.~A. Nesnas, and J.~W. Burdick, ``Motion planning on steep
  terrain for the tethered axel rover,'' in {\em Proc. IEEE Int. Conf. Robot.
  Autom. (ICRA)}, pp.~4188--4195, IEEE, 2011.

\bibitem{shimada2023path}
R.~Shimada and G.~Ishigami, ``Path planning with cable-obstacles avoidance for
  a tethered mobile robot in unstructured environments,'' in {\em Proc. Int.
  Symp. Space Technol. Sci. (ISTS)}, 2023.

\bibitem{shimada2024tangle}
R.~Shimada and G.~Ishigami, ``Tangle- and contact-free path planning for a
  tethered mobile robot using deep reinforcement learning,'' {\em Front. Robot.
  AI}, vol.~11, p.~1388634, 2024.

\bibitem{mcgarey2016trex}
P.~McGarey, F.~Pomerleau, and T.~D. Barfoot, ``System design of a tethered
  robotic explorer ({TReX}) for {3D} mapping of steep terrain and harsh
  environments,'' in {\em Proc. Int. Conf. Field Serv. Robot. (FSR)},
  pp.~267--281, Springer, 2016.

\bibitem{mcgarey2017tslam}
P.~McGarey, K.~MacTavish, F.~Pomerleau, and T.~D. Barfoot, ``{TSLAM}: Tethered
  simultaneous localization and mapping for mobile robots,'' {\em Int. J.
  Robot. Res.}, vol.~36, no.~12, pp.~1363--1386, 2017.

\bibitem{page2022robustness}
J.~J. Page, L.~K. Treers, S.~J. Jorgensen, R.~S. Fearing, and H.~S. Stuart,
  ``The robustness of tether friction in non-idealized terrains,'' {\em IEEE
  Robot. Autom. Lett.}, vol.~8, no.~1, pp.~424--431, 2022.

\end{thebibliography}
\bibliographystyle{ieeetr}

\end{document}